\documentclass{ieee}

\usepackage{times}
\usepackage{graphicx}
\usepackage{subfigure}

\begin{document}

\title{Image decomposition with anisotropic diffusion applied to leaf-texture analysis}

\author{Bruno Brandoli Machado, Wesley Nunes Gon\c{c}alves, Odemir Martinez Bruno \\
Physics Institute of S\~ao Carlos (IFSC) \\
University of S\~ao Paulo (USP)\\
Av. Trabalhador S\~ao-carlense, 400 \\ 
Cx. Postal 369 - S\~ao Carlos - SP - Brasil \\
wnunes@ursa.ifsc.usp.br, brandoli@icmc.usp.br, bruno@ifsc.usp.br \\
}

\maketitle
\thispagestyle{empty}

\begin{abstract}
Texture analysis is an important field of investigation that has received a great deal of interest from computer vision community. In this paper, we propose a novel approach for texture modeling based on partial differential equation (PDE). Each image $f$ is decomposed into a family of derived sub-images. $f$ is split into the $u$ component, obtained with anisotropic diffusion, and the $v$ component which is calculated by the difference between the original image and the $u$ component. After enhancing the texture attribute $v$ of the image, Gabor features are computed as descriptors. We validate the proposed approach on two texture datasets with high variability. We also evaluate our approach on an important real-world application: leaf-texture analysis. Experimental results indicate that our approach can be used to produce higher classification rates and can be successfully employed for different texture applications.
\end{abstract}

\section{Introduction}

Texture plays an important role in pattern recognition and computer vision. Applications with textures are found in several areas, including \textit{remote sensing} \cite{chenPR2008} and \textit{plant leaf identification} \cite{backesIJPRAI2009}. Though texture is easily perceived by humans, it has no precise definition due to its spatial distribution. In addition, physical surface properties produce distinct texture patterns. Thus the lack of a formal definition of texture is reflected into different methods for texture analysis.

Many methods for texture description have been proposed in the literature \cite{zhangPR2002}. They are based on \textit{statistical analysis} of the spatial distribution (e.g., co-occurrence matrices \cite{haralickTSMC1973,haralickIEEE1979} and local binary pattern \cite{kashyapPAMI1986}), \textit{stochastic models} (e.g., Markov random fields \cite{crossPAMI1983}), \textit{spectral analysis} (e.g., Fourier descriptors \cite{azencottPAMI1997}, Gabor filters \cite{gaborJIEE1946} and wavelets transform \cite{daubechies1992}), \textit{structural models} (e.g., mathematical morphology \cite{serra1983} and geometrical analysis \cite{chen1994}), \textit{complexity analysis} (e.g., fractal dimension \cite{mandelbrot1983,odemirIS2008,BackesICIAP2009}), \textit{agent-based model} (e.g., deterministic tourist walk \cite{backesPR2010}). Despite there are effective texture methods, few papers are concerned in enhancing the richness of the texture attribute before computing features.

Inspired by biological vision studies, the community of computer vision has also shown a great deal in representing images using multiple scales. The basic idea is to decompose the original image into a family of derived images \cite{linderbergECSE2008,witkinIJCAI1983}. The decomposition is obtained by convolving the original image with an image operator, for example, a simple way is to employ Gaussian kernels. Although the Gaussian filtering satisfies the heat equation, its derivatives cause spatial distortion in region boundaries. It implies that the diffusion process is equally in all directions, that is, the diffusion is linear or isotropic. On the other hand, Perona and Malik formulate a new concept that modified the linear scale-space paradigm to smooth within a region while preserving edges.

Due to the increasing interest in image analysis, we propose a novel framework to model textures. In the proposed approach, image decomposition using anisotropic diffusion of Perona and Malik is performed before feature extraction. The anisotropic diffusion process is mathematically modeled by partial differential equations (PDEs). The decomposition is applied to extract the texture component, obtained by the difference between the original image and cartoon approximations. Then, Gabor filters are used to extract features from the texture component, which presents more enhanced structures. 

The remaining of this paper is organized as follows. Section \ref{sec:background} presents background information on nonlinear diffusion and Gabor filters. Section \ref{sec:approach} details our approach in texture analysis. Section \ref{sec:exps} presents the results of the experiments performed on two benchmark texture datasets. Finally, conclusions and directions for future research are given in Section \ref{sec:conclusions}.

%-------------------------------------------------------------------------
\section{Background}
\label{sec:background}

In general, texture analysis is studied into five groups: (1) \textit{synthesis}, (2) \textit{segmentation}, (3) \textit{shape from texture}, (4) \textit{compression} and (5) \textit{classification}. All groups have been influenced by the use of decomposition and filter banks. Next we describe image decomposition using anisotropic diffusion and Gabor filters.

\subsection{Anisotropic Diffusion}
\label{sec:ad}

Scale-space theory has been investigated for representing image structures at multiple scales. The idea is to decompose the initial image into a family of derived images. According to \cite{witkinIJCAI1983} and \cite{koenderinkBC1984}, a family of derived images may be viewed as the solution of the heat equation and described using partial differential equations (PDEs). The successful use of PDEs in image analysis is assigned to the power to model many dynamic phenomenon, including diffusion. A new paradigm of nonlinear PDEs for image enhancement was introduced by
Perona and Malik \cite{peronaPAMI1990}. Their formulation, called anisotropic diffusion, uses a nonlinear scheme that smoothes images by creating cartoon approximations, while the region boundaries remain sharp. Formally, the discrete formulation of Perona-Malik is defined as:

\begin{equation}
\label{eq:conv}
I^{t+1}_{i,j} = I^{t}_{i,j} + \left[c_{N}.\nabla_{N}I + c_{S}.\nabla_{S}I + c_{E}.\nabla_{E}I + c_{W}.\nabla_{W}I\right]_{i,j}^{t}
\end{equation}

\noindent where $0 \leq \lambda \leq 1/4$ is a scalar that controls the numerical stability, $\nabla I$ is the gradient magnitude, $c$ is a constant value for the conduction coefficient, $N, S, E$ and $W$ are the mnemonic subscripts for North, South, East and West. The PDE equation above can be write as follows $((i,j) \equiv s)$:

\begin{equation}
\label{eq:conv}
I^{t+1}_{s} = I^{t}_{s} + \frac{\lambda}{\xi_{s}} \sum_{\rho\in\xi_{s}} g(\nabla I_{s,\rho}) \nabla I_{s,\rho}
\end{equation}

\noindent where $I_{s}^{t}$ is the cartoon approximation image, $t$ denotes the number of iterations, $s$ denotes the pixel position, $\xi_s$ represents the number of neighbors of pixel $s$ (usually 4-connectivity), and $g(\nabla I)$ is the conduction function. The value of the gradient is computed by linearly approximating its norm in a specific direction as:

\begin{equation}
\nabla I_{s,\rho} = I_\rho - I_{s}^{t}, \rho \in \xi_s 
\end{equation}

Perona and Malik proposed two functions of diffusion:

\begin{equation}
    g\left(||\nabla I||\right) = e^{-\left(||\nabla I|| / K\right)^2} 
\end{equation}
\noindent and
\begin{equation}
    g\left(|| \nabla I|| \right) = \frac{1}{1 + \left(\frac{||\nabla I||}{K}\right)^2} 
\end{equation}

The parameter $K$ controls the conduction. The first equation favours high contrast edges over low contrast ones, while the latter favours wide regions over smaller ones. Although Perona and Malik proposed two different functions, the smoothed images are quite similar.

\begin{figure*}[!htb]
\centering
 \subfigure[$I_{0}$]{\includegraphics[scale=0.52]{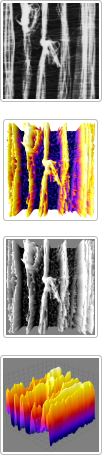}}
 \subfigure[$t_{1}$]{\includegraphics[scale=0.52]{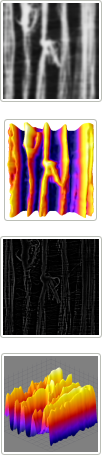}}
 \subfigure[$t_{2}$]{\includegraphics[scale=0.52]{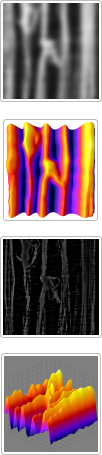}}
 \subfigure[$t_{3}$]{\includegraphics[scale=0.52]{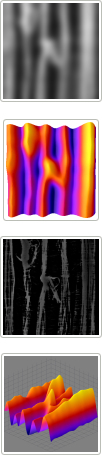}}
 \subfigure[$t_{4}$]{\includegraphics[scale=0.52]{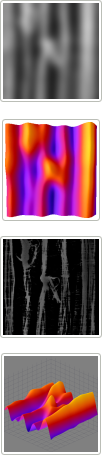}}
 \subfigure[$t_{5}$]{\includegraphics[scale=0.52]{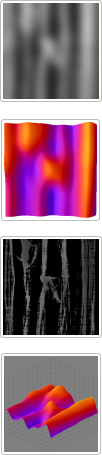}}
 \caption{\label{pic:heat} The essential idea with a scale-space representation of a image is to create a family of cartoon approximations. This figure shows an initial image $I_{0}$ (a) that has been successively smoothed with anisotropic diffusion [(b)--(f)]. The family of derived images may be viewed as the solution of the heat conduction, depicted in rows 2 and 4. The third row corresponds to the texture component.}
\end{figure*}

A texture decomposition with anisotropic diffusion is shown in Figure \ref{pic:heat}. The first row shows the family of cartoon approximations from the original image $I_{0}$. We can observe that the information is gradually smoothed, while textures, in third row, are enhanced by the difference between the original image and cartoon approximations. The solution of the heat diffusion is depicted in rows 2 and 4. Note that the distribution of heat correponds to gray values in the image and the diffusion time is represented by the number of iterations $t$. For different scales $t$ we obtain different levels of smoothing, as shown from $t_1$ (Figure \ref{pic:heat}(b)) to $t_5$ (Figure \ref{pic:heat}(f)).

\subsection{Gabor Filters}

A Gabor filter is a signal sinusoidal plane wave modulated by a Gaussian~\cite{gaborJIEE1946}. The filters used in image decomposition are created from a ``mother'' Gabor function of two dimensions, for a given space $g(x,y)$ and frequency $G(x,y)$ domains. Given the ``mother'' function, a bank of Gabor filters can be obtained in the $g(x,y)$ space domain from operations of dilatations and rotations.

Initially, the Gabor technique generates a filter bank $g_{mn}(x,y)$ for different scales $m=1,\ldots,K$ and orientations $n=1,\ldots,S$ parameters. Texture features are computed by convolving  the original image $I$  with the Gabor filter bank, as depicted in Equation~(\ref{eq:conv}). By tunning the values of $m$ and $n$, some aspects of the image's underlying texture structure can be captured. In this work, a number of 40 Gabor features have been computed (8 orientations and 5 scales).

\begin{equation}
\label{eq:conv}
c_{mn}(x,y) = I(x,y)*g_{mn}(x,y)
\end{equation}

The feature vector $\psi = [E_{11},E_{12},\ldots,E_{KS}]$ is finally obtained by computing the energy of the filtered images according to the Equation~(\ref{eq:ener}).

\begin{equation}
\label{eq:ener}
E_{mn} = \sum_{x,y} [c_{mn}(x,y)]^{2}
\end{equation}

\section{An Approach to Texture Analysis}
\label{sec:approach}

\begin{figure*}[!htb]
\centering
 \subfigure[Original $(f = u+v)$]{\includegraphics[width=0.28\textwidth]{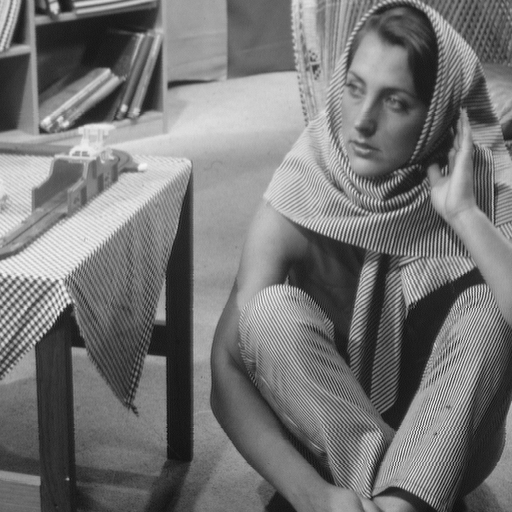}}
 \subfigure[Cartoon $(u)$]{\includegraphics[width=0.28\textwidth]{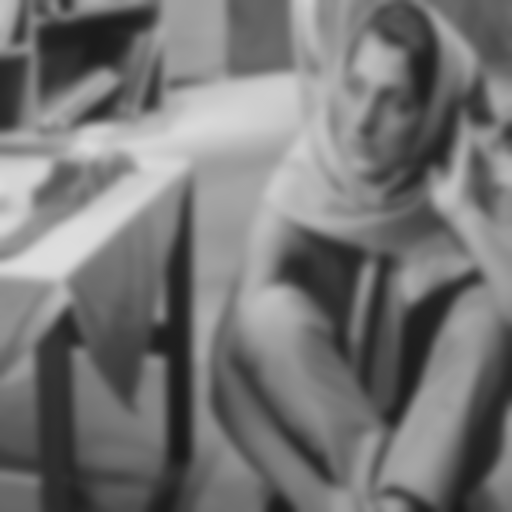}}
 \subfigure[Texture $(v)$]{\includegraphics[width=0.28\textwidth]{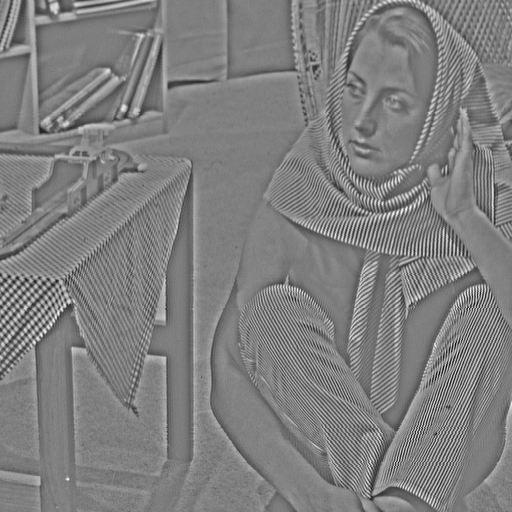}}
 \caption{\label{pic:barbara} An example of image decomposition for the Barbara image (a). At each level of decomposition, it is generated a cartoon approximation $u$ and a texture component $v$. $v$ is obtained by subtracting the original image and the cartoon approximation.}
\end{figure*}

A widely strategy used to compute texture features with Gabor is to construct a bank of filters with different scales and orientations parameters. For each Gabor space is extracted statistical measures, such as energy and entropy. Instead of obtaining right the Gabor space, an original image $(f)$ is decomposed in a set of derived images with anisotropic diffusion of Perona and Malik, described in Section \ref{sec:ad}. This procedure is executed with several levels of decomposition $(t)$ in order to evidence high frequencies $(v)$, while it preserves important structures such as edges. At each level $(t)$, we obtain two components: cartoon approximation $(u)$ and texture $(v)$. The texture component is achieved by subtracting the original image and the cartoon approximation. An example of image decomposition using anisotropic diffusion is shonw in Figure \ref{pic:barbara}.

\begin{figure}[!htb]
	\centering
		\includegraphics[width=1\columnwidth]{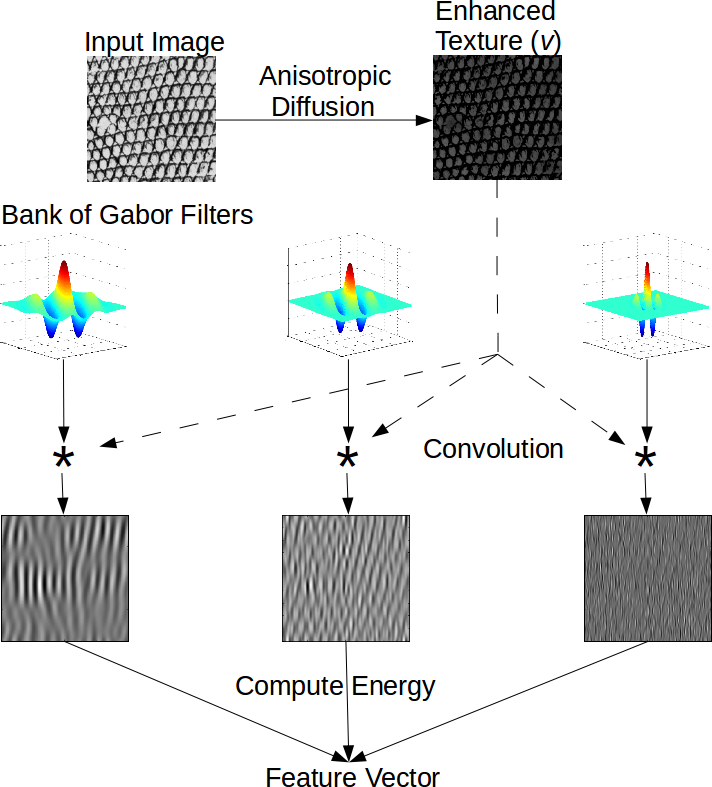}
		\caption{\label{pic:approach} Our approach for texture analysis.}
\end{figure}

The filtering process aim at evidencing high frequencies in the image in order to produce richer representations. Perona and Malik filtering overcomes the main restriction imposed by linear approaches, i.e., blur in region boundaries does not occur. The set of texture images $v$ is then used to extract Gabor features and useful for a variety of tasks, for example, texture classification. The diagram of Figure~\ref{pic:approach} summarizes the approach proposed here.

\section{Experimental Evaluation}
\label{sec:exps}

In order to evaluate our approach, experiments are performed on two image datasets. First, the datasets used
for evaluation are described. Then, implementation details of the descriptors and classifiers are discussed. Finally, the results are shown.

\subsection{Datasets}

\textbf{The Brodatz album} \cite{brodatz1966} is the most known benchmark for evaluating texture methods. Each class is composed by one image divided into nine new samples non-overlapped. These images have $200 \times 200$ pixels with 256 gray levels. A total of 100 texture classes with 10 images per class was used. Recently, the Brodatz dataset has been criticized for certain weaknesses, including lack of viewpoint and scale variation, and illumination changes. Thus, we also use the Vistex dataset. 

\textbf{The Vision Texture dataset} \cite{vistexMIT1995} (or Vistex) contains a large set of natural colorful textures taken under several scale and illumination conditions. In addition, images are acquired with different cameras. For this dataset we use a total of 50 texture classes in gray scale. The size of the original images was $512 \times 512$, but we use the same number of samples as \cite{maenpaaPR2004}. Each texture were split into $128 \times 128$ pixel images, with 16 sub-samples per class, totalizing 800 images.

\subsection{Performance Evaluation}

In the experiments, we compute the energy of Gabor filters with 8 orientations and 5 scales, resulting a feature vector with 40 dimensions. We adopt the K nearest-neighbor (K-NN) classifier, since it is a good reference classification method in the texture recognition. A initial value of $K = 5$ is used, with 10-fold cross validation and Euclidean similarity measure. Here, we change the levels of decomposition $t$ of the anisotropic diffusion process (\textit{scales}). The decomposition ranges from $10$ to $200$. The approach is evaluated using two texture datasets.

\begin{figure}[!htb]
\centering
    \includegraphics[width=1\columnwidth]{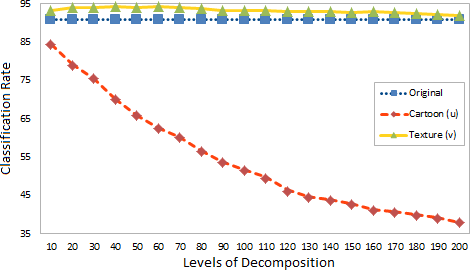}
 \caption{\label{pic:brodatz-exp-it} Comparison of different scales on the Brodatz dataset.}
\end{figure}

\medskip\noindent\textbf{Experiment 1:} First, we evaluate our approach on the Brodatz dataset and compare it to the original Gabor features. Features are computed with different levels of decomposition $t$. Figure~\ref{pic:brodatz-exp-it} shows the classification rates in the $y$ axis, while the levels of decomposition are indicated in the $x$ axis. It can be observed that enhanced texture component $(v)$, extracted using our approach, performs better than the original Gabor method. The highest classification rate ($t= 40$) is $94.29\%$ for texture $(v)$ and $91\%$ for the original Gabor, respectively. Note that the performance of the cartoon approximations $(u)$ get worst at each level of decomposition, which confirms our hypothesis that the component $u$ can be discarded in order to improve the classification rate.

\begin{figure}[!htb]
\centering
    \includegraphics[width=1\columnwidth]{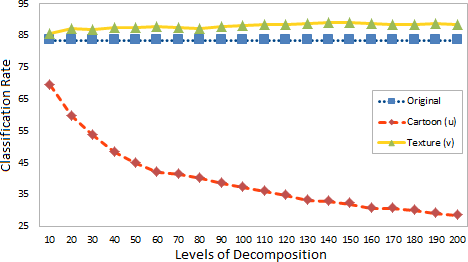}
 \caption{\label{pic:vistex-exp-it} Comparison of different scales on the Vistex dataset.}
\end{figure}

\medskip\noindent\textbf{Experiment 2:} In this experiment we evaluate our approach on the Vistex dataset. The setting for this experiment is the same as the previous one. In Figure \ref{pic:vistex-exp-it}, the classification rates are presented in the $y$ axis, while the decompositions are presented in the $x$ axis. Our approach achieves the best performance with $88.96\%$ ($t=140$) against $83.66\%$ for original Gabor. It is worth noting that the classification rates for the cartoon $(u)$ component reduce at each iteration. This is associated to the gradual decomposition on the image.

Table \ref{table:datasets} presents the average and standard deviation in terms of classification rates. It also shows results for $K = \{3,5,7\}$ on the original image $(f)$, cartoon approximation $(u)$ and texture $(v)$. As we can see, our approach using the texture component $(v)$ outperforms the others for all values of $K$ on both datasets. Interesting results came out from the cartoon approximation experiments, which is discarded in the proposed approach. A classification rate of $67.80\%$ and $31.16\%$ are obtained on the Brodatz dataset and Vistex dataset, respectively. It clearly shows the poor classification power using the cartoon approximation.

\begin{table}[!htb]
 \centering
		 \footnotesize\begin{tabular}{l|l|c|c|c|c}
		 \hline
\textbf{Dataset} & Component & \%(3-NN) & \%(5-NN) & \%(7-NN) \\
		 \hline
\textbf{Brodatz} & $(f)$ Original & 92.53 & 91.00 & 89.04 \\
		 & $(u)$ Cartoon  & 71.61 & 70.06 & 68.96 \\
		 & $(v)$ Texture  & 94.88 & 94.29 & 92.87 \\ 
\cline{2-5}	
	         & $(v-f)$ & 2.35 & 3.29 & \textbf{3.83} \\
		\hline
\textbf{Vistex} & $(f)$ Original & 84.71 & 83.66 & 83.10 \\
		& $(u)$ Cartoon  & 31.85 & 32.95 & 32.35 \\
		& $(v)$ Texture  & 89.21 & 88.96 & 86.65 \\
\cline{2-5}
	        & $(v-f)$ & 4.50 & \textbf{5.30} & 3.55 \\
		\hline
		\end{tabular}
\caption{\label{table:datasets} Comparison of different values of nearest neighbors on both datasets.}
\end{table}

To illustrate the potential of our approach, we compare it with three representative operators used for filtering edges: \textit{Gaussian}, \textit{Laplacian} and \textit{Laplacian of Gaussian} (LoG) (we refer to \cite{forsyth2003} for more details). For all operators, the same procedure of the proposed approach was performed. In this setting, our approach achieved the highest classification rates for all values of $K$ on both datasets. For the Brodatz dataset, an improvement of $3.14\%$ compared to the Gaussian operator was obtained using $K=5$. On the Vistex dataset with $K=5$, our approach achieved a classification rate of $88.96\%$, which is significantly better than the classification rate of $83.63\%$ achieved by the LoG operator. Experimental results demonstrate that our approach is an effective representation for texture modeling.

\begin{table}[!htb]
 \centering
		 \footnotesize\begin{tabular}{l|l|c|c|c|c}
		 \hline
\textbf{Dataset} & Operator + Gabor & \%(3-NN) & \%(5-NN) & \%(7-NN) \\
		 \hline
\textbf{Brodatz} & Gaussian & 92.55 & 91.15 & 89.77 \\
		 & Laplacian & 91.17 & 89.49 & 87.93 \\
		 & LoG & 92.78 & 90.42 & 89.45 \\
		& \textbf{Our approach} & \textbf{94.88} & \textbf{94.29} & \textbf{92.87} \\  
		\hline
\textbf{Vistex} & Gaussian & 85.14 & 82.75 & 81.72 \\
		& Laplacian & 84.56 & 82.71 & 81.05 \\
		& LoG & 85.24 & 83.63 & 82.11 \\
		& \textbf{Our approach} & \textbf{89.21} & \textbf{88.96} & \textbf{86.65} \\ 
		\hline
		\end{tabular}
\caption{\label{table:operators} Comparison of different image operators on both datasets.}
\end{table}

\section{Leaf-Texture Enhancement: A Case Study}
\label{sec:apps}

\begin{figure}[!htb]
\centering
 \subfigure{\includegraphics[width=0.18\columnwidth]{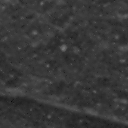}}
 \subfigure{\includegraphics[width=0.18\columnwidth]{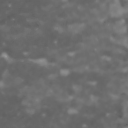}}
 \subfigure{\includegraphics[width=0.18\columnwidth]{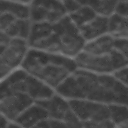}}
 \subfigure{\includegraphics[width=0.18\columnwidth]{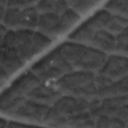}}
 \subfigure{\includegraphics[width=0.18\columnwidth]{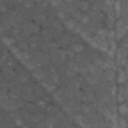}}
 \caption{\label{pic:leaves} Leaf samples.}
\end{figure}

Although there exist some tools interested in identifying plant species, amost none of them are concerned in enhancing the texture attribute before computing features from images. Here, we show a case study using a subset of five classes, with 10 images per class. One example of each class is shown in Figure \ref{pic:leaves}. Again our approach achieved highest classification rates, according to Table \ref{table:leaf}. The results show that our approach is consistent, being a useful method to enhance the texture attribute employed in real-world applications.

\begin{table}[!htb]
 \centering
		 \begin{tabular}{l|c|c|c|c}
		 \hline
Operator + Gabor & \%(3-NN) & \%(5-NN) & \%(7-NN) \\
		 \hline
Original & 76.60 & 75.60 & 74.80 \\
Gaussian & 81.40 & 73.65 & 72.40 \\
Laplacian & 73.80 & 71.80 & 70.80 \\
LoG & 74.20 & 73.00 & 71.40 \\
\textbf{Our Approach} & \textbf{86.00} & \textbf{80.60} & \textbf{76.20} \\
		\hline
		\end{tabular}
\caption{\label{table:leaf} Comparison of different values of nearest neighbors on the leaf dataset.}
\end{table}

\section{Conclusions}
\label{sec:conclusions}

This paper proposed a new approach to enhance the richness of the texture attribute by applying anisotropic diffusion as an early step in the texture image modeling.
We have also demonstrated how the Gabor process can be improved by using our approach.
Promising results have been obtained on two databases of high complexity.
In the Brodatz dataset, experimental results indicate that the proposed approach improves classification rate from $89.04\%$ to $92.87\%$ over the traditional approach.
In addition, experimental results on Vistex dataset demonstrated that the proposed approach provides an improvement of  $5.30\%$ on classification rate.
Our approach is able to successfully handle a wide range of texture methods, e.g. from Gabor filters to Markov random fields. In order to evaluate our approach, we performed it to enhance leaf-texture textures wide used in systems of plant leaf identification.
As part of the future work, we plan to focus on investigating new nonlinear PDEs and texture image methods.

\subsubsection*{Acknowledgments.} The authors gratefully acknowledge the financial support of CNPq and FAPESP.

%-------------------------------------------------------------------------
% \nocite{ref1,ref2}
% \newpage
%\bibliographystyle{ieee}
%\bibliography{acivs}

\end{document}